\pdfoutput=1

\documentclass[11pt]{article}

\usepackage{naacl2021}

\usepackage{times}
\usepackage{latexsym}
\usepackage{graphicx}
\usepackage{amssymb}
\usepackage{amsmath}
\usepackage{hyphenat}
\usepackage{microtype}
\usepackage{multirow}
\usepackage{siunitx}
\usepackage{etoolbox}
\usepackage{algorithm2e}

\DeclareMathOperator*{\argmax}{arg\,max}

\newcommand\irgrretriever{\operatorname{IRGR-retriever}}
\newcommand\irgrgenerator{\operatorname{IRGR-generator}}

\usepackage[T1]{fontenc}

\usepackage[utf8]{inputenc}

\usepackage{microtype}

\usepackage{booktabs}
\usepackage{siunitx} 
\title{Entailment Tree Explanations via \\ Iterative Retrieval-Generation Reasoner}

\author{
Danilo Ribeiro$^{1,2,}$\thanks{~~Work done during an internship at the AWS AI. Code is publicly available at
\url{https://github.com/amazon-research/irgr}.}
~Shen Wang$^2$,
Xiaofei Ma$^2$,
Rui Dong$^2$,
Xiaokai Wei$^2$,
Henry Zhu$^2$,\\
\textbf{Xinchi Chen}$^2$,
\textbf{Zhiheng Huang}$^2$,
\textbf{Peng Xu}$^2$,
\textbf{Andrew Arnold}$^2$,
\textbf{Dan Roth}$^2$,
\\
$^1$Northwestern University,
$^2$AWS AI Labs
\\
\{dnribeiro\}@u.northwestern.edu,\\
\{shenwa, xiaofeim, ruidong, xiaokaiw, henghui\}@amazon.com,\\
\{xcc, zhiheng, pengx, anarnld, drot\}@amazon.com
}

\begin{document}
\maketitle
\begin{abstract}

Large language models have achieved high performance on various question answering (QA) benchmarks, but the explainability of their output remains elusive. Structured explanations, called \textit{entailment trees}, were recently suggested as a way to explain and inspect a QA system's answer. In order to better generate such entailment trees, we propose an architecture called \textit{Iterative Retrieval-Generation Reasoner} (IRGR). Our model is able to explain a given hypothesis by systematically generating a step-by-step explanation from textual premises. The IRGR model iteratively searches for suitable premises, constructing a single entailment step at a time. Contrary to previous approaches, our method combines generation steps and retrieval of premises, allowing the model to leverage intermediate conclusions, and mitigating the input size limit of baseline encoder-decoder models. We conduct experiments using the \textsc{EntailmentBank} dataset, where we outperform existing benchmarks on premise retrieval and entailment tree generation, with around 300\% gain in overall correctness.

\end{abstract}

\section{Introduction}

Large neural network models have successfully been applied to different natural language tasks, achieving state-of-the-art results in many natural language benchmarks. Despite this success, these results came with the expense of AI systems becoming less interpretable \citep{Jain2019, rajani-etal-2019-explain}. 

\begin{figure}
  \includegraphics[width=\columnwidth]{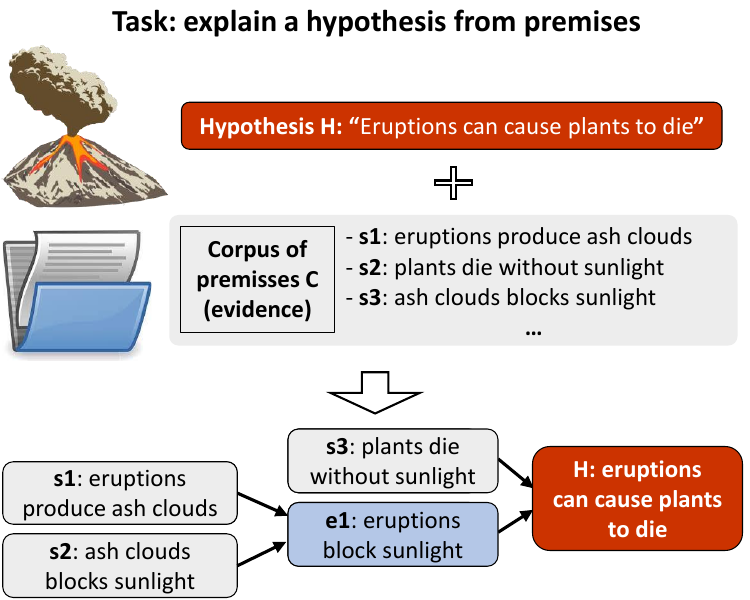}
  \caption{\label{fig:intro}
    Task has as input a hypothesis H (e.g. an answer to a question) and a corpus of premises C (simple textual evidences), the goal is to generate an \textit{entailment tree} that explains the hypothesis H by using premises from C.
  }
\end{figure}

With the desire to make the output of such models less opaque, we propose a question answering (QA) system that is able to explain their decisions not only by retrieving supporting textual evidence (rationales), but by showing how the answer to a question can be systematically proven from simpler textual premises (natural language reasoning). These explanations are represented using \textit{entailment trees}, as depicted in Figure \ref{fig:intro}. First introduced by \citet{dalvi2021explaining}, entailment trees represents a chain of reasoning that shows \textit{how} a hypothesis (or an answer to a question) can be explained from simpler textual evidence. In comparison, other explanation approaches such as retrieval of passages (rationales) \citep{deyoung-etal-2020-eraser} or multi-hop reasoning (chaining) \citep{jhamtani2020grc} are less expressive than entailment trees, which are comprised of multi-premise textual entailment steps.

\begin{figure*}
  \centering
  \includegraphics[width=4.2in]{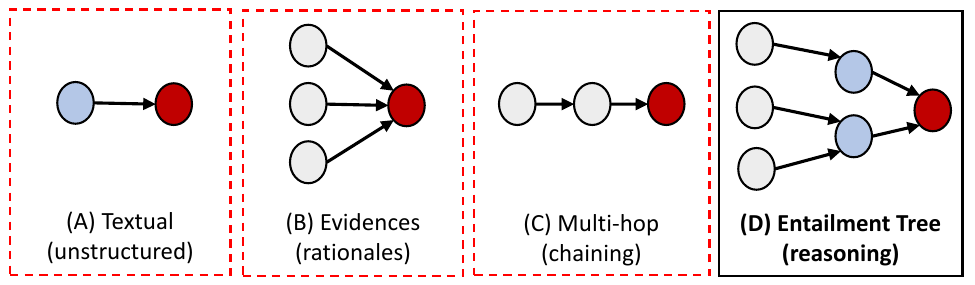}
  \caption{\label{fig:explanation_comparison} Comparison among different natural language explanation approaches. The images show (A) plain textual explanations (B) retrieval of evidence passages (C) multi-hop explanations (D) entailment trees. The approach in (D) allows for more detailed inspection of the reasoning behind an explanation. Nodes in gray are retrieved from a corpus, nodes in blue are generated, and the red node is the hypothesis or answer that is being explained.}
\end{figure*}

In order to generate such entailment trees, previous works \citep{tafjord-etal-2021-proofwriter, dalvi2021explaining, bostrom-etal-2021-flexible} have used encoder-decoder models that takes as input a small set of retrieved premises and output a linearized representation of the entailment tree. Such models are limited by (1) the language model’s fixed input size, and they may construct incorrect proofs when the retrieval module cannot fetch all relevant premises at once (2) such approaches do not leverage the partially generated entailment trees. In contrast, we propose \textit{\textbf{I}terative \textbf{R}etrieval-\textbf{G}eneration \textbf{R}easoner} (IRGR), a novel architecture that iteratively searches for suitable premises, constructing a single entailment step at a time. At every generation step, the model searches for a distinct set of premises that will support the generation of a single step, therefore mitigating the language model’s input size limit and improving generation correctness.

Our contributions are two-fold. First, we design a retrieval method that is able to better identify premises needed to generate a chain of reasoning, which explains a given hypothesis. Our retrieval method outperforms previous baselines by 48.3\%, while allowing for a dynamic set of premises to be retrieved. Secondly, we propose an iterative retrieval-generation architecture that constructs partial proofs and augments the retrieval probes using intermediate generation results. We show that integrating the retrieval module with iterative  generation can significantly improve explanations. Our proposed approach achieves new state-of-the-art result on entailment tree generation with over 306\% better results on the All-Correct metric (strict comparison with golden data), while using a model with one order of magnitude fewer parameters. 

\section{Related Work}

\begin{figure*}
  \includegraphics[width=\textwidth]{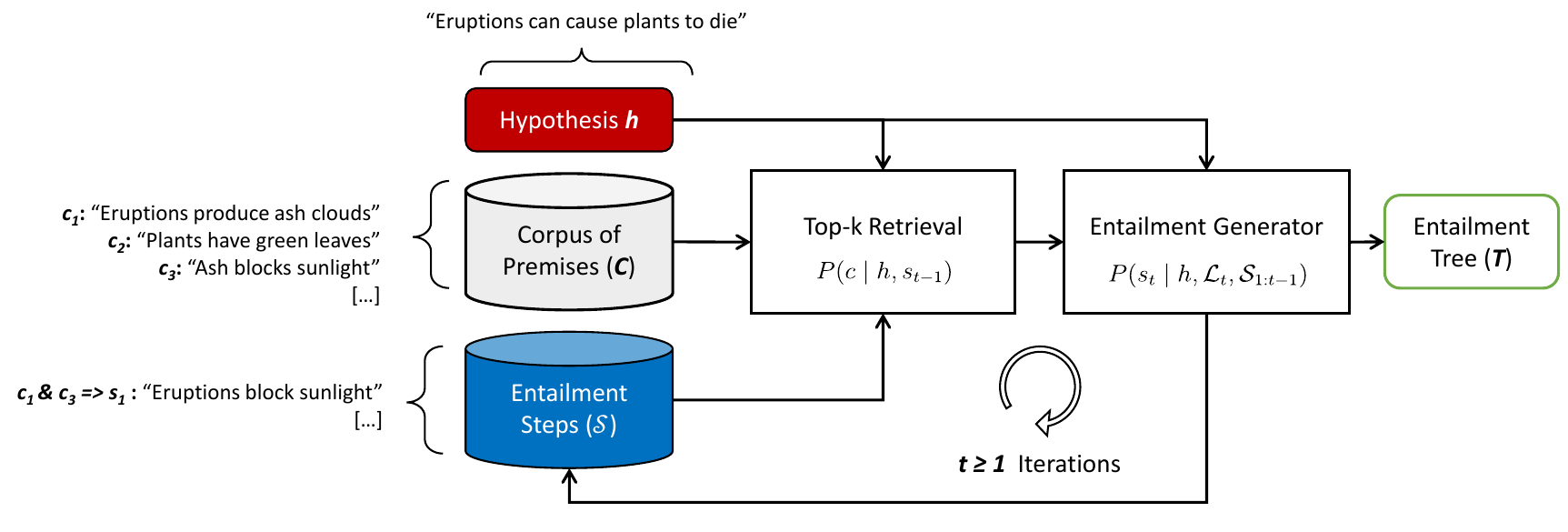}
  \caption{\label{fig:overview} IRGR is composed of two modules, $\irgrretriever{}$ and $\irgrgenerator{}$. The $\irgrretriever{}$ iteratively fetches a set of premises from a corpus $C$ in order to generate an entailment tree (structured explanation for a given hypothesis). The $\irgrgenerator{}$ computes a single entailment step at a time, and the intermediate generated steps are stored and used for subsequent retrieval and generation. }
\end{figure*}

Traditionally, natural language processing (NLP) frameworks were based on white-box methods such as rule-based systems \citep{allen1988natural, ribeiro2019predicting, ribeiro2021combining} and decision trees \citep{boros-etal-2017-fast}, which were inherently inspectable \citep{danilevsky-etal-2020-survey}. More recently, large deep learning language models (black-box methods) have gained popularity \citep{song2020mpnet, DBLP:journals/jmlr/RaffelSRLNMZLL20}, but their improvements in result quality came with a cost: the system's outputs lack explainability and inspectability.

There have been many attempts to mitigate this issue, including input perturbation \citep{ribeiro2018anchors} and premises selection \citep{deyoung-etal-2020-eraser}. One promising explanation approach is to combine the model's output with a human-interpretable explanation. For instance, \citet{NIPS2018_8163} introduced the concept of natural language explanation in their e-SNLI dataset while \citet{rajani2019explain} expanded this idea to commonsense explanations. \citet{jhamtani2020grc} further explored the notion of explanation in multi-hop QA, where explanations contain a \textit{chain of reasoning}, instead of simple textual explanations. Different from these explanation approaches, our work generates explanations in the form of \textit{entailment trees}, introduced by \citet{dalvi2021explaining}, which are composed of multi-premise textual entailment steps. Entailment trees are more detailed explanations, making it easier to inspect the reasoning behind the model's answer. Figure \ref{fig:explanation_comparison} shows a diagram compering different natural language explanation methods according to their structure and use of textual evidence.

The first approach used to generate entailment trees was based on the \textit{EntailmentWriter} model by \citet{tafjord-etal-2021-proofwriter}. However, their approach is limited by the input size of the encoder-decoder language models, where a fixed set of supporting facts is used to generate an explanation. Instead, our model iteratively fetches a set of premises using dense retrieval conditioned on previous entailment steps, allowing for more precise explanations.

Our work is also related to some recent approaches that combine retrieval and neural networks for QA tasks \citep{karpukhin-etal-2020-dense, guu2020realm}. The work of \citet{NEURIPS2020_6b493230} combined dense retrieval with encoder-decoder models, where a different set of passages were retrieved for each generated character. Conditioning the retrieval of a passage on previously retrieved passages has been explored in the context of multi-hop QA \citep{zhao2021beamdr, xiong2021answering}, and multi-hop explanations \citep{valentino2021hybrid, cartuyvels-etal-2020-autoregressive}. However, these approaches either are not used to generate explanations or do not use inferred intermediate reasoning steps to retrieve premises.


\section{Approach}

\subsection{Problem Definition}

The problem input consists of a \textbf{corpus of premises} $C$ (simple textual statements) and a \textbf{hypothesis} $h$. The objective is to generate an \textbf{entailment tree} $T$ that explains the hypothesis $h$ by using a subset of the premises in $C$ as building blocks. Entailment trees are represented as a tuple $T = (h, \mathcal{L}, \mathcal{E}, \mathcal{S})$, where \textbf{leaf nodes} $l_i \in \mathcal{L}$ are retrieved from the corpus (i.e. $\mathcal{L} \subseteq C$), \textbf{internal tree nodes} $e_i \in \mathcal{E}$ are intermediate conclusions (new sentences not present in corpus $C$, note that intermediate conclusions are generated by the model), and $s_i \in \mathcal{S}$ is a list of \textbf{entailment steps} that can explain the hypothesis $h$, which is always the tree root and the final conclusion. An illustration of the problem and expected entailment tree can be found in Figure \ref{fig:intro}. 

Each entailment step $s_i$ represents one inference step from a conjunction of premises to a conclusion. For instance, ``$l_1 \wedge l_2 \Rightarrow e_1$'' or ``$l_1 \wedge l_2 \wedge e_1  \Rightarrow h$'' could be valid entailment steps in $\mathcal{S}$. Note that the root of $T$ is always the node representing $h$.

\subsection{Architecture}

Our approach, which we call \textit{\textbf{I}terative \textbf{R}etrieval-\textbf{G}eneration \textbf{R}easoner} (IRGR), consists of two modules, the $\irgrretriever{}$ and the $\irgrgenerator{}$. The initial input to the model is the hypothesis $h$ and the corpus of premises $C$. The generation process is performed through multiple iterations. At each iteration step $t \geq 1$ the $\irgrretriever{}$ selects a subset of premises from the corpus $\mathcal{L}_{t} \subseteq C$.  The $\irgrgenerator{}$ outputs one entailment step $s_{t}$ per iteration until the entailment tree $T$ is fully generated. Given $\mathcal{S}_{1:t-1} = (s_1, \dots, s_{t-1})$ as the list of the entailment steps generated up to the previous iterations $t-1$, the generator takes as input $\mathcal{L}_{t}$ and $\mathcal{S}_{1:t-1}$ and produces the next entailment step $s_{t}$. The generation stops when the entailment step's conclusion is the hypothesis $h$, i.e., the proof is finished. Formally, the $t$-th iteration of the generation process is defined as:
\begin{align}
\mathcal{L}_{t} & = \irgrretriever(h, s_{t-1}) \\
s_{t} & = \irgrgenerator(h, \mathcal{L}_{t}, \mathcal{S}_{1:t-1})
\end{align}

The $\irgrretriever$ searches over the premises in corpus $C$ using \textit{dense passage retrieval} \citep{karpukhin-etal-2020-dense}. Meanwhile, the $\irgrgenerator$ was implemented using T5, the Text\hyp{}to\hyp{}Text Transformer \citep{DBLP:journals/jmlr/RaffelSRLNMZLL20}, while any other sequence\hyp{}to\hyp{}sequence language model could also be used. An overview of the model can be seen in Figure \ref{fig:overview}.


\subsubsection{IRGR-retriever}
\label{sec:retrieval-of-premises}

The $\irgrretriever$ module proposed in this work aims to retrieve premises from the corpus $C$. In existing baseline models the retrieval is done in one single step, fetching a fixed set of premises before generation \citep{tafjord-etal-2021-proofwriter}. However, the generation of entailment trees requires a different set of leaves for each entailment step. To address this issue, our $\irgrretriever$ fetches $k_{t}$ premises from $C$ to produce $\mathcal{L}_{t}$ at iteration step $t$. Note that the size of $C$ can be very large ($k_{t} << \lvert C \rvert$). The value $k_{t}$ is chosen such that the size of $\mathcal{L}_{t}$ is small enough to fit in the context of a language model while still being large enough to fetch as many premises as possible (in our experiments, the value $k_t$ is always below $25$). We define the retrieval probability of a premise $c \in C$ at a certain iteration step $t$ as:
\begin{equation}
P(c \mid h, s_{t-1}) = \dfrac{exp(\langle \mathbf{c}, \mathbf{q_t} \rangle)}{ \sum_{c' \in C} exp(\langle \mathbf{c'}, \mathbf{q_t} \rangle)}
\end{equation}


Where $\phi$ is the sentence encoder function used to encode both premises and hypothesis, transforming the input text into a dense vector representation in $\mathbb{R}^M$. The values $\mathbf{c} = \phi(c)$, $\mathbf{c'} = \phi(c')$ and $\mathbf{q_t} = \phi(h, s_{t-1})$ are dense $M$-dimensional vectors. The operator $\langle . \rangle$ represents the inner product between two vectors. 

The encoder follows the Siamese Network architecture from \citet{reimers-gurevych-2019-sentence}. We select a set of $N$ positive and negative examples in the form of query-value pairs $\{(q_j, c_j)\}^N_{j=1}$ for training. Queries $q_j$ encode both the hypothesis $h$ and previous entailment step $s_{t-1}$ by concatenating their textual values. The positive examples are taken from the golden entailment trees, where $c_j \in \mathcal{L}$. For negative examples, we pair a query $q_j$ with either random premises from $C$ or premises retrieved using the not fine-tuned version of the encoder (hard negatives).

We define $\hat{y}_j$ as the label given to the training example $(q_j, c_j)$. For positive examples, the label $\hat{y}_j$ depends on how close the leaf node $l_i \in \mathcal{L}$ is to the intermediate step $s_{t-1}$ in the golden tree:
\begin{equation}
\hat{y}_j = \left\{
  \begin{array}{@{}ll@{}}
    0, & \text{if negative} \\
    \lambda, & \text{if positive and}\ l_i \not\in \texttt{ant}(s_{t-1}) \\ 
    1, & \text{if positive and}\ l_i \in \texttt{ant}(s_{t-1})
  \end{array}\right.
\end{equation}

Where $\texttt{ant}(s_{t-1})$ denotes the set of antecedents in some entailment step $s_{t-1}$, and $l_i \in \texttt{ant}(s_{t-1})$ means that the leaf node $l_i$ is used in the entailment step $s_{t-1}$. The value $\lambda \in [0:1]$ is used to give lower priority to leaf nodes not relevant to the current entailment step ($\lambda$ = 0.75 gave the best results in our experiments). Finally, we fine-tune the encoder $\phi$ by minimizing the following loss function $L_{\phi}$, where $N$ is the number of training examples:
\begin{equation}
L_{\phi} = \frac{1}{N}\sum_{j=1}^{N} \left( \hat{y}_j - \dfrac{\langle \phi(q_j), \phi(c_j) \rangle}{\|\phi(q_j)\| \| \phi(c_j) \|} \right)
\end{equation}

\begin{figure}
  \includegraphics[width=\columnwidth]{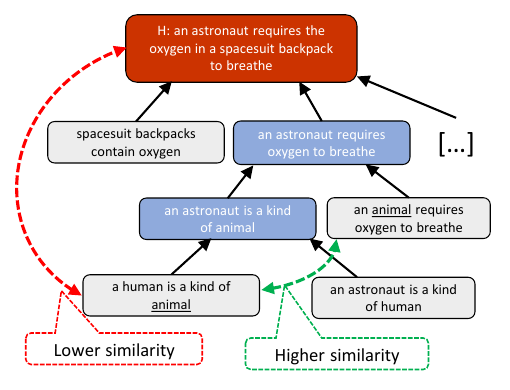}
  \caption{\label{fig:low-similarity-retrieval}
  Entailment tree example showing how some retrieval examples are challenging. Leaf sentences are not always directly related to hypothesis.
  }
\end{figure}

\SetKwComment{Comment}{/* }{ */}
\RestyleAlgo{ruled}
\SetKwInOut{Parameter}{Parameter}

\begin{algorithm}[t]
\caption{Conditional Retrieval}\label{alg:conditioned-retrieval}
\KwData{hypothesis $h$, corpus $C$, number of retrieved premises $k_{0}$}
\KwResult{retrieved premises $\mathcal{L}_{0}$}
\Parameter{conditioning factor $\omega$}
$Q \gets \{ h \}$ \Comment*[r]{set of queries}
$\mathcal{L}_{0} \gets \{ \}$\;
\For{$i\gets0$ \KwTo $k_{0}$}{
    $C' \gets \{c \in C : c \not\in \mathcal{L}_{0} \}$\;
    \eIf{$i \geq \omega$}{
        $l_i \gets \argmax_{(c \in C')} P(c \mid Q)$\;
        $Q \gets Q \cup \{ l_i \}$\;
    }{
        $l_i \gets \argmax_{(c \in C')} P(c \mid h)$\;
    }
    $\mathcal{L}_{0} \gets \mathcal{L}_{0} \cup \{ l_i \}$\;
}
\end{algorithm}

One significant challenge is that for the first generation step, when $t=1$, the list of previously generated entailment steps $\mathcal{S}_{0}$ is empty. The retrieval only depends on $h$, meaning $\mathcal{L}_{1} = \irgrretriever(h)$. It is more difficult to retrieve premises for leaf nodes when the entailment tree's depth is large since the leaf nodes have low syntactic and semantic similarity with the hypothesis $h$. For instance, the example in Figure \ref{fig:low-similarity-retrieval} shows how leaf node ``\textit{a human is a kind of animal}'' (depth 3) is needed to build the entailment tree, but is syntactically distinct to hypothesis ``\textit{an astronaut requires the oxygen in a space suit backpack to breath}''.

To mitigate this problem, we perform a conditional retrieval on the first step, where the retrieval module uses partial results as part of the query, as depicted in Algorithm \ref{alg:conditioned-retrieval}. This algorithm assumes that leaf nodes (premises) further from the root node (hypothesis) are more similar to each other than to the root node itself. The parameter $\omega$ (value $\omega = 15$ yields the best results on development set) is used to split the search, such that part of the retrieved premises only depend on the hypothesis $h$. In contrast, the other parts of the retrieved premises depend on the hypothesis and previously retrieved premises stored in the set $Q$.

\subsubsection{IRGR-generator}

The $\irgrgenerator{}$ consists of a sequence-to-sequence model that outputs one single entailment step given a context. One key aspect of this module is encoding the input and output as plain text. 

\paragraph{Encoding Entailment Trees:} Entailment trees are linearized from leaves to root. Each leaf node $l_i \in \mathcal{L}$, intermediate node $e_i \in \mathcal{E}$ and root node $h$ are encoded with the symbols ``\texttt{sent}'', ``\texttt{int}'' and ``\texttt{hypothesis}'', respectively. The entailment steps represent conjunctions with ``\texttt{\&}'' and entailment with the symbol ``\texttt{->}''. For instance, the entailment tree depicted in Figure \ref{fig:intro} can be represented as:
\begin{quote}
``\texttt{sent1} \texttt{\&} \texttt{sent2} \texttt{->} \texttt{int1:} \texttt{Eruptions} \texttt{block} \texttt{sunlight;} \texttt{sent3} \texttt{\&} \texttt{int1} \texttt{->} \texttt{hypothesis;}''
\end{quote}
Note that the text of intermediate nodes have to be explicitly represented, since they are not part of the corpus $C$. Ultimately, they have to be generated by the model. The input to the model encodes the hypothesis $h$ and retrieved premises $l^{t}_{i} \in \mathcal{L}_{t}$, which are straightforwardly encoded as follows:
\begin{quote}
``\texttt{hypothesis:} \texttt{Eruptions} \texttt{can} \texttt{cause} \texttt{plants} \texttt{to} \texttt{die;}\\ \texttt{sent1:} \texttt{eruptions} \texttt{emit} \texttt{lava}\\ \texttt{sent2:} \texttt{eruptions} \texttt{produce} \texttt{ash} \texttt{clouds} \\\texttt{sent3:} \texttt{[...];}''
\end{quote}

When a leaf sentence $l^{i}_{t}$ is used in the entailment step, it is removed from the context for the following step, and the premise \texttt{sent} identifier is not used to encode new retrieved premises. A detailed example of input and output for the $\irgrgenerator$ module is shown in Appendix \ref{sec:generator-output-examples}.

\section{Experiments}

\subsection{Datasets}

We evaluate our architecture on the \textsc{EntailmentBank} dataset \citep{dalvi2021explaining}, which is comprised of 1,840 questions (each associated with a hypothesis $h_i$ and entailment tree $T_i$) with 5,881 total entailment steps. On average, each entailment tree has 7.6 nodes (including leaf, intermediate, and root) and around 3.2 entailment steps. The corpus of premises $C$ has around 11K entries and is derived from the WorldTree V2 \citep{xie-etal-2020-worldtree} in addition to a few premises created by the EntailmentBank annotators.

\subsection{Evaluation Metrics}

\begin{table}[t]
\centering
\sisetup{ table-align-uncertainty=true,
            separate-uncertainty=true,
            detect-weight=true,
            detect-inline-weight=math  }
\begin{tabular}{lSS}
\toprule
 \multicolumn{1}{l}{\textbf{Method}} & \multicolumn{1}{c}{\textbf{R@25}} & \multicolumn{1}{c}{\textbf{All-Correct}}  \\
\midrule
Okapi BM25  & \multicolumn{1}{c}{45.01} & \multicolumn{1}{c}{22.35}  \\
EntailmentWriter  & \multicolumn{1}{c}{59.76} & \multicolumn{1}{c}{34.70}  \\
\midrule
IRGR-retriever (sing.) & \multicolumn{1}{c}{64.41} & \multicolumn{1}{c}{40.29}   \\
IRGR-retriever (cond.) &  \multicolumn{1}{c}{\bfseries 68.28} & \multicolumn{1}{c}{44.70}   \\

IRGR-retriever* & \multicolumn{1}{c}{-} & \multicolumn{1}{c}{\bfseries 51.47}   \\ 
\bottomrule
\end{tabular}
\caption{\label{tab:retrieval-results}
Retrieval results. The methods with * retrieves more than 25 premises from corpus.}
\end{table}

\begin{table*}[t]
\centering
\sisetup{ table-align-uncertainty=true,
            separate-uncertainty=true,
            detect-weight=true,
            detect-inline-weight=math  }
\begin{tabular}{l|S[table-format=2.1, round-precision=1,detect-weight,mode=text] S[table-format=2.1, round-precision=1,detect-weight,mode=text] S[table-format=2.1, round-precision=1,detect-weight,mode=text] S[table-format=2.1, round-precision=1,detect-weight,mode=text] S[table-format=2.1, round-precision=1,detect-weight,mode=text] S[table-format=2.1, round-precision=1,detect-weight,mode=text] S[table-format=2.1, round-precision=1,detect-weight,mode=text]}
\toprule
\multirow{2}{*}{\textbf{Method}} & \multicolumn{2}{c}{\textbf{Leaves}} & \multicolumn{2}{c}{\textbf{Steps}} & \multicolumn{2}{c}{\textbf{Intermediates}} & \multicolumn{1}{c}{\textbf{Overall}} \\
 & \textbf{F1} & \textbf{All-Cor.}  & \textbf{F1} & \textbf{All-Cor.} & \textbf{F1} & \textbf{All-Cor.} & \textbf{All-Cor.} \\
\midrule
 EntailmentWriter  $\dagger$  & 39.9 & 3.8 & 7.4 & 2.9 & 35.9 & 7.1 & 2.9 \\
 IRGR (Ours) & 45.63 & \multicolumn{1}{S[table-format=2.1, round-precision=1,detect-weight,mode=text]}{\bfseries 11.8} & \multicolumn{1}{S[table-format=2.1, round-precision=1,detect-weight,mode=text]}{\bfseries 16.1} & \multicolumn{1}{S[table-format=2.1, round-precision=1,detect-weight,mode=text]}{\bfseries 11.4} & \multicolumn{1}{S[table-format=2.1, round-precision=1,detect-weight,mode=text]}{\bfseries 38.8} & \multicolumn{1}{S[table-format=2.1, round-precision=1,detect-weight,mode=text]}{\bfseries 20.9} & \multicolumn{1}{S[table-format=2.1, round-precision=1,detect-weight,mode=text]}{\bfseries 11.5} \\
 - w/o iter. & \multicolumn{1}{S[table-format=2.1, round-precision=1,detect-weight,mode=text]}{\bfseries 46.6} & 10.0 & 11.29 & 8.24 & 38.72 & \multicolumn{1}{S[table-format=2.1, round-precision=1,detect-weight,mode=text]}{\bfseries 20.9} & 8.24 \\
 - w/o iter. \& cond. & 36.14 & 3.82 & 6.06 & 3.24 & 30.53 & 10.29 & 3.24 \\

\bottomrule
\end{tabular}
\caption{\label{tab:table-generation-scores} 
Entailment tree scores for baseline methods and IRGR, along four different dimensions (test set). F1 scores measure predicted/gold overlap, while All-Correct scores are 1 when all the predictions for a tree are correct, 0 otherwise. $\dagger$ indicates results using T5-11B model.}
\end{table*}

\subsubsection{Retrieval}

We evaluate our $\irgrretriever$ module using two different sets of metrics. The first one is ``Recall at k'' (R@k), a standard evaluation metric for information retrieval. The second metric ``All-Correct'' is more strict, and the results are only considered correct if all the premises from the golden tree are retrieved. Formally, given the retrieved premises $\mathcal{L}$ and the set of gold premises $\mathcal{L}^{*}$, the metrics R@k is given by $\lvert \mathcal{L} \cap \mathcal{L}^{*} \rvert \mathbin{/} \lvert \mathcal{L}^{*} \rvert$, and the metric All-Correct is 1 if $\lvert \{ x \in \mathcal{L}^{*} : x  \not\in \mathcal{L} \} \rvert = 0$, or 0 otherwise. For our experiments, we consider k = 25 since that's roughly the maximum number of sentences that can fit in the T5 language model's 512 tokens context.

\subsubsection{Entailment Tree Generation}

We adopt the evaluation metrics defined by \citet{dalvi2021explaining}, which compares the generated entailment tree $T = (h, \mathcal{L}, \mathcal{E}, \mathcal{S})$ with the golden entailment tree $T^{*} = (h, \mathcal{L}^{*}, \mathcal{E}^{*}, \mathcal{S}^{*})$. The metrics evaluate the correctness along four dimensions: (1) leaf nodes,  (2) entailment steps, (3) generated intermediate nodes, (4) and overall correctness. The first step is to align the nodes from $T$ with the nodes from $T^{*}$ by Jaccard similarity (alignment algorithm and further details of metrics described in Appendix \ref{sec:metrics-details}). This method tries to ignore variations between predicted and gold trees that do not change the semantics of the output. The four metric dimensions are described below as follows. For each metric with F1 value, there is also a strict ``All-Correct'' metric that is equal to 1 when F1 = 1 and 0 otherwise.

\paragraph{Leaf (F1, All-Correct):} Tests if the predicted and golden leaf nodes match. This metric compares the sets $\mathcal{L}$ and $\mathcal{L}^{*}$ using F1 score.

\paragraph{Steps (F1, All-Correct):} Tests if the predicted entailment steps follow the correct structure. Given that $s_i \in \mathcal{S}$ matches $s_j \in \mathcal{S}^{*}$ according to the alignment algorithm, tests if the premises of $s_i$ are equal to those of $s_j$, and computes the F1 score according to the set of all matched steps.

\paragraph{Intermediates (F1, All-Correct):} Tests if the sentences of the generated intermediate nodes are correct. Given that intermediate nodes $e_i \in \mathcal{E}$ and $e_j \in \mathcal{E}^{*}$ were matched by the alignment algorithm, the F1 score is computed by comparing the textual similarity between the set of the aligned and correct pairs $e_i$ and $e_j$.

\paragraph{Overall (All-Correct):} Tests all previous metrics together. The All-Correct value is only 1 if the All-Correct values for leaves, steps, and intermediates are 1. Note that this is a strict metric, and any semantic difference between $T$ and $T^{*}$ will cause the score to be zero.

\subsection{Implementation Details}

All experiments were conducted using a machine with 4 Tesla V100 GPUs with 16GB of memory. Our code is based on HuggingFace's Transformers \cite{wolf2019huggingface} implementation of the \texttt{t5-large} model \citep{DBLP:journals/jmlr/RaffelSRLNMZLL20}. The retrieval module uses the Sentence Transformers \citep{reimers-gurevych-2019-sentence} sentence embeddings by fine-tuning the \texttt{all-mpnet-base-v2} encoder. Please refer to Appendix \ref{sec:appendix-experiment-details} for further details on hyper-parameters and training settings.

\begin{table*}[t]
\centering
\begin{tabular}{ll|SSSSSSS}
\toprule
 \multirow{2}{*}{\textbf{Task}} & \multirow{2}{*}{\textbf{Method}} & \multicolumn{2}{c}{\textbf{Leaves}} & \multicolumn{2}{c}{\textbf{Steps}} & \multicolumn{2}{c}{\textbf{Intermediates}} & \multicolumn{1}{c}{\textbf{Overall}} \\
 & & \textbf{F1} & \textbf{All-Cor.}  & \textbf{F1} & \textbf{All-Cor.} & \textbf{F1} & \textbf{All-Cor.} & \textbf{All-Cor.} \\
\midrule
\multirow{2}{*}{\textbf{Gold}} & EntailmentWriter & \multicolumn{1}{c}{98.7} & \multicolumn{1}{c}{86.2} & \multicolumn{1}{c}{50.5} & \multicolumn{1}{c}{37.7} & \multicolumn{1}{c}{67.6} & \multicolumn{1}{c}{36.2} & \multicolumn{1}{c}{33.5} \\
& IRGR (Ours) & \multicolumn{1}{c}{99.6} & \multicolumn{1}{c}{97.6} & \multicolumn{1}{c}{51.1} & \multicolumn{1}{c}{37.6} & \multicolumn{1}{c}{66.8} & \multicolumn{1}{c}{34.1} & \multicolumn{1}{c}{32.1} \\
\midrule
\multirow{2}{*}{\textbf{Gold+Dist.}} & EntailmentWriter & \multicolumn{1}{c}{84.3} & \multicolumn{1}{c}{35.6} & \multicolumn{1}{c}{35.5} & \multicolumn{1}{c}{22.9} & \multicolumn{1}{c}{61.8} & \multicolumn{1}{c}{28.5} & \multicolumn{1}{c}{20.9} \\

& IRGR (Ours) & \multicolumn{1}{c}{69.9} & \multicolumn{1}{c}{23.8} & \multicolumn{1}{c}{30.5} & \multicolumn{1}{c}{22.3} & \multicolumn{1}{c}{47.7} & \multicolumn{1}{c}{26.5} & \multicolumn{1}{c}{21.8}   \\
  

\bottomrule
\end{tabular}
\caption{\label{tab:table-generation-scores-wo-retrieval}
Entailment tree scores for baseline methods and IRGR, along four different dimensions (test set). The ``Gold'' and ``Gold+Dist.'' tasks  do not require retrieval and evaluates solely on the model's entailment tree generation capabilities.}
\end{table*}

\subsection{Results}

\subsubsection{Retrieval Results}


We compare our retrieval module against two baselines: \textbf{Okapi BM25} and the retrieval module of \textbf{EntailmentWriter}, which constitutes of a classifier that retrieves relevant sentences using RoBERTA \citep{liu2020roberta} and performs re-ranking with Tensorflow-Ranking-BERT \citep{DBLP:journals/corr/abs-2004-08476}. 

For comparison, we break down the results of our approach (the $\irgrretriever$ module) into three variations. The \textbf{IRGR-retriever (sing.)} method retrieves premises from the corpus using a single query element, namely the hypothesis $h$. The \textbf{IRGR-retriever (cond.)} method performs conditioned retrieval as described by Algorithm \ref{alg:conditioned-retrieval}. This retrieval method is not iterative and fetches a fixed set of premises per example. Finally, \textbf{IRGR-retriever} tries to emulate the retrieval when combined with the generation module. It not only performs conditional retrieval, but also fetches a different set of premises for each iteration depending on the generated intermediate nodes. In this retrieval experiment, the IRGR-retriever uses the intermediate nodes from the golden entailment trees. Therefore, IRGR-retriever results should be considered an upper bound since the generator might not produce the desirable intermediate steps used for queries.

Table \ref{tab:retrieval-results} shows the R@25 and All-Correct metrics results for different methods. Our premise retrieval module performs consistently better than baselines. For instance, the ``IRGR-retriever (cond.)'' outperforms the retriever from EntailmentWriter by 14.2\% on R@25 and 28.8\% on All-Correct metric. Note that ``IRGR-retriever'' may retrieve a variable number of premises (greater than 25), so we are not reporting R@25 for this method.

\subsubsection{Entailment Tree Generation Results}

We compare our method against EntailmentWriter baseline model on entailment tree generation\footnote{Note that the results are updated from NAACL 2022 camera ready version.}. As shown in Table \ref{tab:table-generation-scores}, our method outperforms the EntailmentWriter in all metrics. The overall tree structure better matches the golden tree, where the score for Overall All-Correct metric has an impressive increase of over 300.0\%. Note that EntailmentWriter uses the T5-11B model, which has around 10 times more parameters than our model.

We also show the ablation results of combining different retrieval modules with our proposed generation module on Table \ref{tab:table-generation-scores}. The ``\textbf{w/o iter.}'' method does not iteratively retrieve premises, relying on one-shot retrieval at the beginning of the generation. As for the ``\textbf{w/o iter. \& cond.}'' method, the model does not use the conditioned retrieval, only relying on the trained dense retrieval with the hypothesis $h$ as the query instead. 

The work of \citet{dalvi2021explaining} defines two other simplified entailment tree generation tasks for further ablation studies. We report the results for what they define as ``Task-1'' and ``Task-2'', which are generation tasks where the golden premises are given as input, disregarding the retrieval component. Results in Table \ref{tab:table-generation-scores} report what they define as ``Task-3''. For clarity, we rename ``Task-1'' and ``Task-2'' to ``Gold'' and ``Gold+Dist.'', respectively, and show the results in Table \ref{tab:table-generation-scores-wo-retrieval}. In the ``Gold'' task, each context uses the golden leaves as input, while the ``Gold+Dist.'' task uses the golden leaves plus some distractors (up to 25 distractors). When comparing models with the same number of parameters (we use their reported T5-large results), the generation results without retrieval are roughly the same as the EntailmentWriter method. This experiment shows that the iterative generation can create accurate explanations compared to a single pass generation when using golden retrieved premises.


\subsection{Results Breakdown}

\begin{figure}
  \includegraphics[width=\columnwidth]{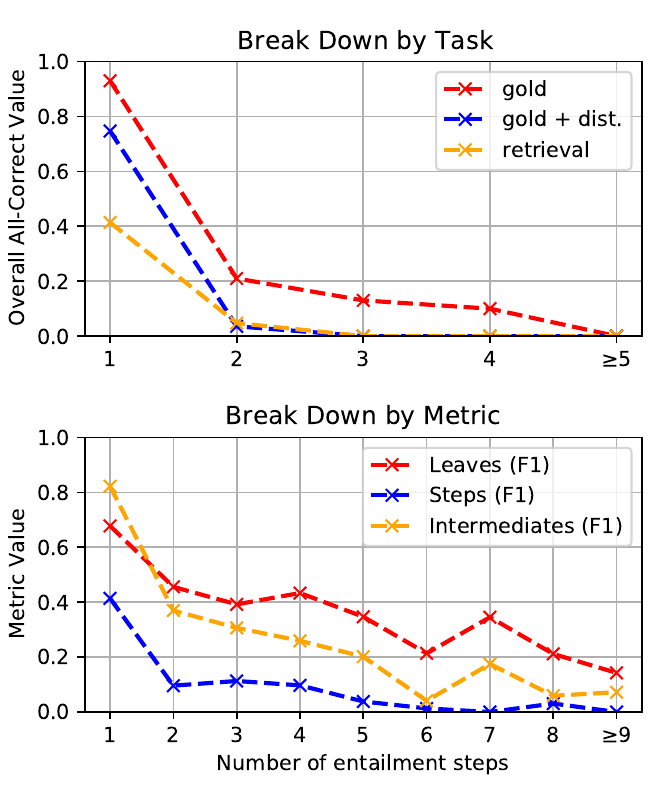}
  \caption{\label{fig:result-break-down}
    Result breakdown for number of steps in explanation (entailment steps). 
  }
\end{figure}

We investigate how well the system performs relative to the number of steps in the gold tree. Figure \ref{fig:result-break-down} contains two graphs with results breakdown. The graph on the top shows the all-correct metric values for all three tasks (golden, golden + distractors, and retrieval). The bottom graph shows all F1 metrics (leaves, steps, and intermediates), but only for the ``retrieval'' task.

The results demonstrate that generating entailment trees becomes increasingly difficult as the size of the tree increases. The IRGR model cannot perfectly predict trees with more than four steps for any of the three different tasks. For the ``retrieval'' task (without the golden leaf sentences provided as input), the IRGR model cannot successfully generate trees with three or more steps. This could be explained by the fact that the all-correct metric is very strict, and missing or misplacing a single leaf sentence can result in an incorrect tree. 

This downwards trend is also present in the ``Break Down by Metrics'' graph. Most noticeably, the ``Steps (F1)'' metric is especially challenging, having values close to zero for entailment trees with more than five steps. This metric is one of the main bottlenecks that lowers the value of the ``Overall All-Correct'' metric.

\subsection{Analysis}

To understand the strengths and weaknesses of our model, we conduct further analysis of the output of the IRGR. When analyzing errors in the generation of entailment trees, we use the results on the development set for the task with distractors. We manually annotate 50 predicted trees that contain some error compared to the golden tree. We categorize the different types of errors, identifying both individual generated steps errors and entailment tree errors.


\subsubsection{Retrieval Error Analysis}

We use \textsc{EntailmentBank}'s development set to automatically compute metrics that will give us some insights into the type of errors made by the $\irgrretriever$ module. We use ``IRGR-retriever (cond.)'' to fetch a set of 25 premises for each data point, where we identify the set of \textbf{true positives} (correctly retrieved premises) and the set of \textbf{false negatives} (missing premises).

To understand if the false negatives are more challenging to retrieve than the true positives, we compute the number of overlapping uni-grams and bi-grams between premises and hypotheses in these two sets. We notice that true positives contain $28.5\%$ more uni-gram overlap and $68.6\%$ more bi-gram overlap to the hypothesis compared to the false negatives. These results suggest that premises lexically \textit{dissimilar} to the hypothesis are, in theory, more challenging to retrieve.
    
We also investigate how the depth (number of edges in a path from the tree root) of a leaf node in the gold tree correlates to the errors of the $\irgrretriever$ module. We compute the average depth of true positive nodes as $2.3$, while for false-negative nodes, the average depth is $3.0$. These results strengthen the idea that leaf nodes deeper in the tree tend to be harder to retrieve, as depicted in Figure \ref{fig:low-similarity-retrieval}.

\subsubsection{Entailment Step Error Analysis}

The first error case is called \textbf{invalid entailment steps} (56\% of errors), meaning that the conclusion of a step did not follow from the premises. For instance, in ``\textit{kilogram is used to measure heavy objects}'' $\wedge$ ``\textit{an automobile is usually a heavy object}'' $\Rightarrow$ ``\textit{kilogram can be used to measure the mass of an automobile}'', the model assumes that ``measure'' is the same as ``measure of mass'', even though that is not explicitly stated. 

The second error case accounts for \textbf{misevaluation and irrelevance} (27\% of errors). It happens when the step is correct but does not match the golden tree, or when the step is correct but is not relevant or well placed in the final entailment tree. In the third error case, labeled \textbf{repetition} (17\% of errors), the conclusion directly copied the premises, not creating a new sentence for the intermediate step.

\subsubsection{Entailment Tree Error Analysis}

When analyzing errors between the entire generated and golden trees, we noticed that \textbf{incorrect or missing leaves} (52\% of errors) is the most common type of problem. For instance, when explaining the hypothesis ``\textit{light year can be used to measure the distance between the stars in milky way}'' the premises ``\textit{the milky way is a kind of galaxy}'' and ``\textit{a galaxy is made of stars}'' are missing from the generated tree, making it impossible to explain the second part of the hypothesis.

The remaining errors are categorized as \textbf{invalid or skipped steps} (32\% of errors), where the model commonly concludes an invalid conclusion from premises. This error often overlaps with \textbf{missing leaves} due to the fact that the model uses fewer premises when it skips important intermediate steps; \textbf{Imperfect evaluation} (12\% of errors), where the tree produced is valid, but does not match the golden tree; \textbf{Disconnected or degenerate trees} (4\% of errors), where the generated output does not form a tree, or follows the desired output format.

\section{Conclusion}

As deep learning models become more ubiquitous in the natural language field, it is desirable that users can understand the model's answer by inspecting the reasoning chain from simple premises to the answer hypothesis. To generate rich, systematic explanations, we proposed a method that can iteratively generate and retrieve premises to produce entailment trees. We show how our approach has advantages over previous baselines, where the retrieved premises and generated explanations are more accurate.

In future work, we plan to improve the generation module by leveraging the structure of the entailment tree instead of relying purely on the encoder-decoder models. This idea could potentially fix the issues with ``invalid entailment steps'' and ``repetition'', which account for 73\% of entailment step errors. We also plan to understand how explanations can be generated in the case of a false hypothesis, where we would expect the model to build a conclusion explaining why a statement is incorrect. It could help users verify false claims and understand the meaning behind their incorrectness.


\section*{Acknowledgements}
We are thankful to Felicity M. Lu-Hill for proofreading this paper. The research  leading to this paper was supported in part by the Machine Learning, Reasoning, and Intelligence Program of the Office of Naval Research.

\bibliography{anthology,custom}
\bibliographystyle{acl_natbib}

\appendix

\section{Appendix}
\label{sec:appendix}

\subsection{Experiment Details}
\label{sec:appendix-experiment-details}


The $\irgrgenerator$ used the \texttt{T5-large}\footnote{Model available in https://huggingface.co/t5-large} model from HuggingFace library. The best models were chosen according to the best ``Overrall All-Correct'' metric on the validation set. During training, we used the following hyper parameters: learning rate: $3 \cdot 10^{-5}$, epochs: $15$, training batch size: $4$, validation batch size: $4$, max number of input tokens: $512$, max number of output tokens: $256$, warm-up steps: $0$, weight decay: $0$.

The $\irgrretriever$ module uses the version
\texttt{all-mpnet-base-v2}\footnote{Model available in https://huggingface.co/sentence-transformers/all-mpnet-base-v2}  from the SentenceTransformers library. During training, we used the following hyper parameters: learning rate: $5 \cdot 10^{-5}$, epochs: $10$, training batch size: $32$, validation batch size: $32$, loss function: \textit{cosine similarity loss}, warm-up steps: $0$, weight decay: $0$. 


\subsection{Entailment Tree Evaluation Metrics Details}
\label{sec:metrics-details}

The alignment algorithm between the nodes from gold and predicted entailment trees can be roughly described as follows:

\begin{enumerate}
  \item For each intermediate node $e_i \in \mathcal{E}$ and $e_j \in \mathcal{E}^{*}$, compute the set of leaf sentences in which the nodes are ancestors.
  \item Align each intermediate node $e_i$ to the first golden node $e_j$ for which the Jaccard similarity of their respective set of associated leaf sentences is maximum. If any node $e_i$ is associated with no gold nodes $e_j$ (Jaccard similarity is zero), then $e_i$ is aligned to a blank node (no conclusion).
\end{enumerate}

Given the list of aligned nodes, the metrics ``Intermediates (F1,  All-Correct)'' compute the similarity between two generated intermediate steps using $BLEURT$ \citep{sellam-etal-2020-bleurt}, a learned evaluation metric based on BERT. We use the BLEURT-Large-512 model to compute textual similarity scores. The prediction between intermediate nodes is considered correct if $BLEURT(e_i, e_j) > 0.28$ (this threshold was originally picked by \citet{dalvi2021explaining} using a subset of 300 manually labeled pairs).

\subsection{Generator Input and Output Examples}
\label{sec:generator-output-examples}

The generation is done in multiple steps. Below are the input and output examples for a tree with two entailment steps (T=1 and T=2). In the following example, only the golden premises are used, which is how the generator is trained. At test time this format is extended to use the retrieved premises instead.

\paragraph{INPUT T=1:} ``\texttt{hypothesis:} \texttt{notebook} \texttt{paper} \texttt{can} \texttt{be} \texttt{recycled} \texttt{many} \texttt{times;} \texttt{sent1:} \texttt{recyclable} \texttt{means} \texttt{a} \texttt{material} \texttt{can} \texttt{be} \texttt{recycled} \texttt{/} \texttt{reused} \texttt{many} \texttt{times} \texttt{sent2:} \texttt{paper} \texttt{is} \texttt{recyclable} \texttt{sent3:} \texttt{notebook} \texttt{paper} \texttt{is} \texttt{a} \texttt{kind} \texttt{of} \texttt{paper;}''

\paragraph{OUTPUT T=1:} ``\texttt{sent2} \texttt{\&} \texttt{sent3} \texttt{->} \texttt{int1:} \texttt{notebook} \texttt{paper} \texttt{is} \texttt{recyclable;}''

\paragraph{INPUT T=2:} ``\texttt{hypothesis:} \texttt{notebook} \texttt{paper} \texttt{can} \texttt{be} \texttt{recycled} \texttt{many} \texttt{times;} \texttt{sent1:} \texttt{recyclable} \texttt{means} \texttt{a} \texttt{material} \texttt{can} \texttt{be} \texttt{recycled} \texttt{/} \texttt{reused} \texttt{many} \texttt{times;} \texttt{sent2} \texttt{\&} \texttt{sent3} \texttt{->} \texttt{int1:} \texttt{notebook} \texttt{paper} \texttt{is} \texttt{recyclable;}''

\paragraph{OUTPUT T=2:} ``\texttt{int1} \texttt{\&} \texttt{sent1} \texttt{->} \texttt{hypothesis;}''

\paragraph{} Note that the input for \textbf{T=2} removed the premises used in the previous entailment step, i.e. ``\texttt{sent2}'' and ``\texttt{sent3}'', and added the generated entailment step from \textbf{T=1} to the end of the input. 

\end{document}